\documentclass[letterpaper]{article} 
\usepackage{aaai2026}  
\usepackage{times}  
\usepackage{helvet}  
\usepackage{courier}  
\usepackage[hyphens]{url}  
\usepackage{graphicx} 
\usepackage{natbib}  
\usepackage{caption} 
\usepackage{algorithm}
\usepackage{algorithmic}

\usepackage{newfloat}
\usepackage{listings}
\DeclareCaptionStyle{ruled}{labelfont=normalfont,labelsep=colon,strut=off} 
\floatstyle{ruled}
\newfloat{listing}{tb}{lst}{}
\floatname{listing}{Listing}
\usepackage{xspace}
\usepackage{amsmath}
\usepackage{cleveref}
\usepackage{booktabs}
\usepackage{multirow}

\usepackage{xcolor}
\usepackage{tcolorbox}

\newcommand{\benchmarkname}{ME2\xspace}
\newcommand{\taskname}{multimodal solution explanation\xspace}

\title{Explain with Visual Keypoints Like a Real Mentor!\\A Benchmark for Multimodal Solution Explanation}
\author{
    Jaewoo Park\textsuperscript{\rm 1}\equalcontrib,
    Jungyang Park\textsuperscript{\rm 1,\rm 2}\equalcontrib,
    Dongju Jang\textsuperscript{\rm 1},
    Jiwan Chung\textsuperscript{\rm 1},\\
    Byungwoo Yoo\textsuperscript{\rm 2},
    Jaewoo Shin\textsuperscript{\rm 2},
    Seonjoon Park\textsuperscript{\rm 2},
    Taehyeong Kim\textsuperscript{\rm 2},
    Youngjae Yu\textsuperscript{\rm 3}
}
\affiliations{
    \textsuperscript{\rm 1}Yonsei University\\
    \textsuperscript{\rm 2}Mathpresso\\
    \textsuperscript{\rm 3}Seoul National University\\
    jerife@yonsei.ac.kr, youngjaeyu@snu.ac.kr
}

\begin{document}

\maketitle  

\begin{abstract}
With the rapid advancement of mathematical reasoning capabilities in Large Language Models (LLMs), AI systems are increasingly being adopted in educational settings to support students' comprehension of problem-solving processes. However, a critical component remains underexplored in current LLM-generated explanations: multimodal explanation. In real-world instructional contexts, human tutors routinely employ visual aids, such as diagrams, markings, and highlights, to enhance conceptual clarity.
To bridge this gap, we introduce the \textit{\taskname} task, designed to evaluate whether models can identify visual keypoints, such as auxiliary lines, points, angles, and generate explanations that incorporate these key elements essential for understanding.
To evaluate model performance on this task, we propose \benchmarkname, a multimodal benchmark consisting of 1,000 math problems annotated with visual keypoints and corresponding explanatory text that references those elements.
Our empirical results show that current models struggle to identify visual keypoints.
In the task of generating keypoint-based explanations, open-source models also face notable difficulties.
This highlights a significant gap in current LLMs' ability to perform mathematical visual grounding, engage in visually grounded reasoning, and provide explanations in educational contexts.
We expect that the \taskname task and the \benchmarkname dataset will catalyze further research on LLMs in education and promote their use as effective, explanation-oriented AI tutors.\end{abstract}

\begin{links}
    \link{Archive}{https://me2-benchmark.github.io}
\end{links}

\begin{figure}[!t]
    \centering
    \includegraphics[width=1\linewidth]{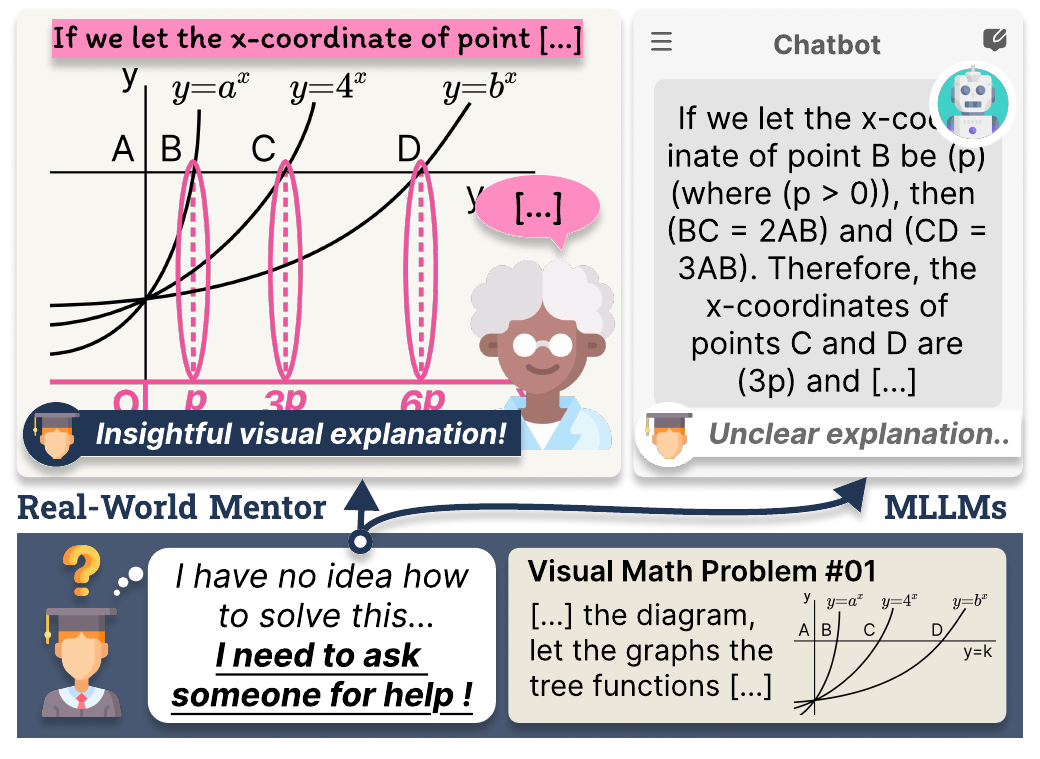}
    \caption{A student solving a math problem often benefits from visual cues—such as lines, symbols, or highlights—that human instructors use to aid understanding, unlike current AI models that focus solely on textual solutions. To serve as effective educational assistants, machines must go beyond answer generation and emulate human-like explanation strategies by explicitly incorporating and referencing visual elements. }
    \label{fig:teaser}
\end{figure}

\section{Introduction}
The traditional one-to-many educational model (i.e., one teacher for multiple students) is gradually transitioning to one-to-one personalized tutoring systems and online learning \cite{mukul2023digital}. Recent developments in Multimodal Large Language Models (MLLMs) have opened new opportunities for effective learning, such as estimating question difficulty \cite{park2024large}, assisting teachers in curriculum planning \cite{hu2024teaching}, and supporting interactive tutoring systems \cite{chevalier2024language}. In particular, numerous studies \cite{liu2023mathematical, uesato2022solving, lu2023mathvista} have focused on enhancing the mathematical reasoning abilities of MLLMs. As a result, MLLMs have led many students to use them as tools when faced with mathematical questions \cite{pardos2024chatgpt}. 

However, from a student’s perspective, relying solely on the reasoning footprints of MLLMs may not always be the best way to understand problems \cite{pardos2024chatgpt, jia2024comparison}. One might wonder what distinguishes a broadly comprehensible explanation from a solution that merely yields the correct answer, for either a human or a model? A critical factor is the use of visual cues.

In actual educational settings, Dual Coding Theory (DCT) naturally occurs, providing effective learning opportunities for students \cite{paivio2013imagery, paivio1990mental}. According to DCT, combining verbal and visual information enhances student comprehension \cite{clark1991dual}. As illustrated in \Cref{fig:teaser}, human mentors often use visual scaffolding, such as annotated diagrams or highlighted keypoints on a blackboard, to foster intuitive understanding \cite{arcavi2003role, stylianou2010teachers, lee2024vistavisualintegratedtailored}. In contrast, current AI models lack the capacity to generate such visual explanations. Moreover, existing datasets~\cite{hendrycks2021measuring, lu2023mathvista, wang2024measuring} focus solely on problem-solving and overlook educational objectives, making them insufficient for developing models capable of providing such forms of multimodal instructional support.

To address these limitations, we introduce \textit{\taskname}, a novel task that aims to enhance models’ capacity to generate educationally effective and visually grounded mathematical explanations.
In this task, models are required to (1) identify visual keypoints that are not present in the original problem but are crucial for understanding (e.g., lines, angles, annotations), and (2) generate explanatory text that explicitly refers to them.
To benchmark performance on \taskname, we propose Multimodal Explanations for Mathematics Education (\textbf{\benchmarkname}) benchmark.
The \benchmarkname includes not only the problem and its solution, but also annotations of the visual keypoints that serve as visual cues necessary to explain the solution, as well as keypoint-based explanatory texts aligned with them.
Notably, we emphasize that \benchmarkname goes beyond simple problem-solving, offering educational value in addressing the previously unexplored dimension of \taskname.

Experiments on \benchmarkname demonstrate that current MLLMs struggle to reliably identify visual keypoints. While closed-source models show potential in generating explanations grounded in visual keypoints, open-source generalist and math-specialized models show limited ability in this aspect. This suggests that current models largely fail to achieve robust mathematical visual grounding and visually grounded reasoning in educational contexts.
We believe that \benchmarkname will catalyze research toward strengthening mathematical visual grounding and reasoning, and advancing models that can serve as effective and student-friendly educational mentors.

Our contributions are as follows:  
\begin{enumerate}  
    \item \textbf{A \taskname task} that supports students' educational comprehension by identifying critical visual keypoints and generating explanatory text that explicitly references them.  
    \item \textbf{A \benchmarkname benchmark}, rooted in authentic educational contexts, to rigorously assess \taskname performance and facilitate further research on model-based explanations in real-world settings. 
    \item \textbf{Extensive experimental evaluations} of state-of-the-art MLLMs on \taskname task, highlighting current limitations in recognizing and leveraging crucial visual keypoints to support effective learning.
\end{enumerate}

\begin{figure*}[!t]
    \includegraphics[width=1\linewidth]{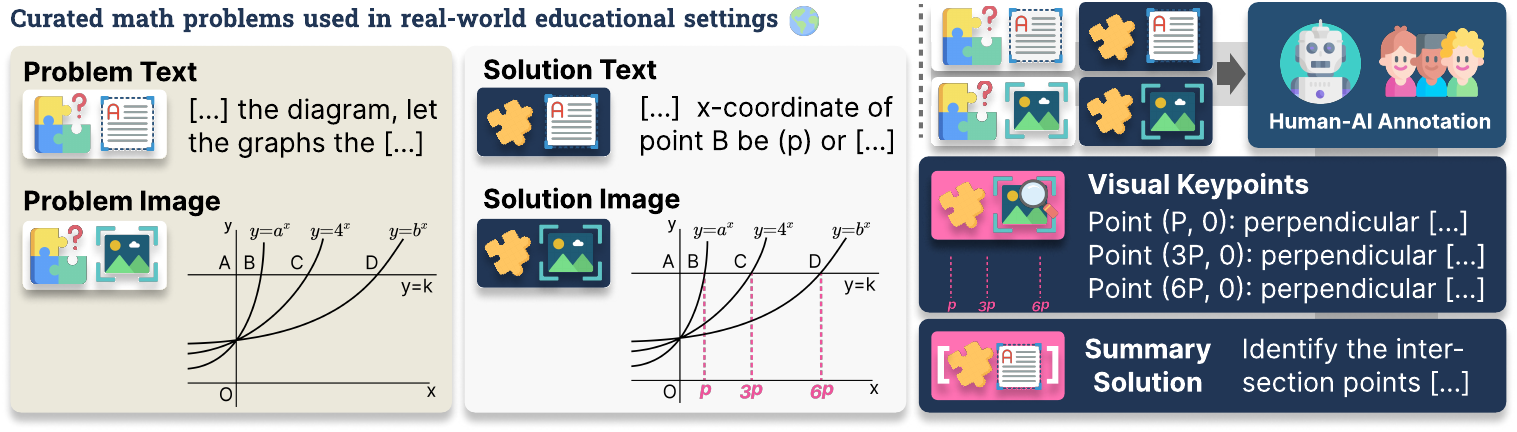}
    \caption{An overview of the \benchmarkname benchmark. The \benchmarkname consists of multimodal problem–solution pairs curated from real-world educational settings, along with visual keypoints and explanation summaries generated through a Human–AI annotation.}
    \label{fig:benchmark_overview}
\end{figure*}

\begin{figure*}[!t]
    \includegraphics[width=1\linewidth]{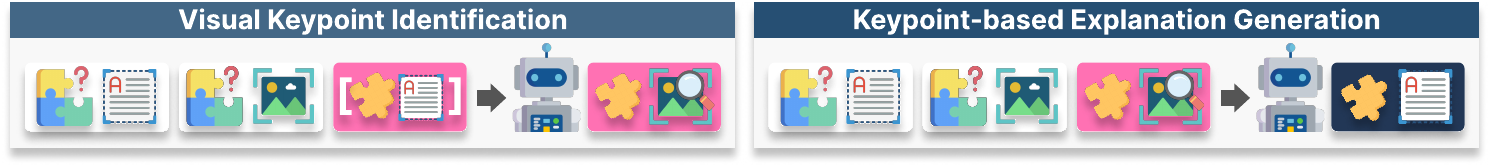}
    \caption{We propose two subtasks to robustly analyze \taskname capacity: (1) Visual Keypoint Identification, which challenges machines to recognize visual keypoints useful for subsequent explanation, and (2) Keypoint-based Explanation Generation, which requires models to generate explanations that explicitly reference the identified visual keypoints.}
    \label{fig:task_overview}
\end{figure*}


\section{Related Works}
\paragraph{Language Models for Education.}
Recent advances in Large Language Models (LLMs) \citep{NEURIPS2020_1457c0d6} have sparked significant interest in educational applications, particularly in personalized problem recommendation \citep{park2024large}, automated tutoring \citep{chevalier2024language}, and the provision of tailored feedback and customized curricula \citep{hu2024teaching, macina2023mathdial, feng2023citing}. For effective education, research suggests that combining textual and visual information enhances comprehension and memory more effectively than using text alone \citep{arcavi2003role, stylianou2010teachers, lee2024vistavisualintegratedtailored}. As \citet{clark1991dual} explains, dual representations supply multiple retrieval cues and cultivate richer mental models. Building on this insight, we introduce the \taskname task to enable LLMs to offer learners more comprehensive learning opportunities. This task enables the model to pinpoint the visual keypoints essential for students’ comprehension and to generate explanations grounded in those keypoints, thereby delivering more comprehensive educational support and ultimately improving the overall quality of their learning experience.

\paragraph{Mathematical Benchmarks.}
Current LLMs show strong performance on mathematical problems, making them valuable tools for students \citep{zhuang2024math, luo2025ursa}. To evaluate these models, traditional mathematical benchmarks~\citep{cobbe2021training, hendrycks2021measuring} have been crucial in assessing reasoning capabilities. With the growing multimodal capabilities of LLMs, benchmarks such as MathVista~\citep{lu2023mathvista}, Math-Vision~\citep{wang2024measuring}, and MathVerse~\citep{zhang2024mathverse} have extended this evaluation to image-based math problems. Recent efforts like OlympiadBench~\citep{he2024olympiadbench} and MM-MATH~\citep{sun2024mm} further assess models not just on final answers but also on their reasoning processes. However, most existing benchmarks focus solely on problem-solving, overlooking educational objectives. To address this gap, we introduce \benchmarkname, which advances beyond problem-solving to evaluate a model’s capacity to generate visually and logically coherent explanations and key visual cues that support effective instructional use.

\section{\benchmarkname Benchmark}
\benchmarkname is a \taskname benchmark consisting of 1,000 instances. Each of which contains a problem text ($T_{p}$), a problem image ($I_{p}$), an explanatory solution text ($T_{s}$), a solution image ($I_{s}$), and visual keypoints ($VK$) that newly highlight elements crucial for understanding (e.g., lines, angles), and a concise explanation summary ($T_{s}^{tldr}$) to anchor the model’s explanatory solution direction.
To create a benchmark that can assess the \taskname capabilities of MLLMs, we define the visual keypoint $VK$ and summary of the explanation $T_{s}^{tldr}$ in Section~\ref{subsec:benchmark_construction}. An overview of \benchmarkname and its construction is illustrated in \Cref{fig:benchmark_overview}.

\subsection{Benchmark Construction}
\label{subsec:benchmark_construction}
\paragraph{In-house Data Curation.}
We extract 1,000 instances of multimodal problem–solution pairs \(\langle T_{p},\,I_{p},\,T_{s},\,I_{s} \rangle\) from an in-house mathematics education platform. All instances were authored by domain experts in mathematics to support effective student learning and are derived from materials authentically used in real-world educational contexts.
The instances are written in Korean and span middle- to high-school levels, with a primary focus on geometry and graph theory. All benchmark data were carefully curated to ensure compliance with copyright regulations.

To benchmark the models' ability to recognize visual keypoints, we ensure that each solution image \(I_{s}\) is derived from the corresponding problem image \(I_{p}\) by adding only new elements such as points, angles, lines, regions, and symbols while preserving the original structure.  
For instance, in \Cref{fig:benchmark_overview}, points are added to problem image \(I_{p}\) to produce solution image \(I_{s}\).  
This setup allows us to accurately evaluate whether a model can identify the critical visual keypoints and effectively incorporate them into its explanation.

We strictly curate the dataset to single-image math problems with either multiple-choice or short-answer formats. We focus on the domains of geometry and graph to ensure that visual context is essential for solving each problem. Each sample in \benchmarkname consists of two natural language texts (the problem text $T_p$ and the solution text $T_s$) and two RGB images (the problem image $I_p$ and the solution image $I_s$).

\paragraph{Annotation Process.}
To create the textual-form visual keypoints, we streamline the simple yet labor-intensive task of comparing problem and solution images by using GPT-4o \cite{achiam2023gpt} as an auxiliary tool. The model produces an initial set of keypoints $\{vk^{\text{ai}}_1, \dots, vk^{\text{ai}}_n\} \in VK^{\text{ai}}$, which four annotators, each holding a bachelor’s degree in science or engineering, verify and refine for precision and consistency with our annotation guidelines:

Any element that is newly added or modified, including points, lines, angles, regions, or symbols such as parallel marks, congruence marks, right-angle marks, or length labels, must be recorded in the format \{\texttt{element: description}\}, where \texttt{element} identifies the visual feature and \texttt{description} explains how it is introduced with reference to surrounding features.

Once the verified keypoints are fixed, human annotators and the AI tool jointly generate a brief, keypoint-aligned summary of each solution text ($T_{s}^{tldr}$). Since explanations may follow multiple valid paths (see Appendix), this summary anchors a single solution direction during model explanation generation, ensuring an unambiguous consensus set of visual keypoints.
Finally, the entire benchmark was translated from Korean to English using an AI tool and then reviewed by two bilingual annotators. From a 10\% subset, annotators achieved substantial agreement, with Cohen’s~$\kappa$ of 0.84~\citep{cohen1960coefficient}, indicating strong reliability.
Consequently, each \benchmarkname\ instance is represented as
$\langle T_{p}, I_{p}, T_{s}, VK, T_{s}^{\text{tldr}}\rangle$.

\begin{table}[t!]
\centering
\begin{tabular}{lr}
\toprule
Total problem–solution pairs      & 1,000 \\
- Geometry                        & 763   \\
\quad - Multiple-choice questions          & 464   \\
\quad - Short-answer questions    & 299   \\
- Graph                        & 237   \\
\quad - Multiple-choice questions & 141   \\
\quad - Short-answer questions    & 96    \\ \midrule
Average number of $VK$            & 3.73   \\
Maximum words in $T_{p}$          & 211   \\
Maximum words in $T_{s}$          & 361   \\
Maximum words in $vk_{n}$         & 45    \\
Maximum words in $T_{s}^{tldr}$   & 198   \\
Average words in $T_{p}$          & 53.1  \\
Average words in $T_{s}$          & 101.4 \\
Average words in $vk_{n}$         & 12.2  \\
Maximum words in $T_{s}^{tldr}$   & 35.8    
\\ \bottomrule
\end{tabular}
\caption{Statistics of the \benchmarkname benchmark, including problem subjects, types, and instance word counts.}
\label{tab:data_analysis}
\end{table}

\subsection{Data Analysis}
\label{subsec:data_analysis} 
The \benchmarkname benchmark consists of 1,000 problem–solution pairs: 763 (76.3\%) geometry problems and 237 (23.7\%) graph problems. Among these, 605 (60.5\%) are multiple-choice questions and 395 (39.5\%) are short-answer questions. It spans 17 chapters (see \Cref{fig:data_analysis}) and 33 sections (see Appendix).
On average, each sample contains about 3.8 visual keypoints $VK$, derived from annotations.
These keypoints fall into four main categories: points, lines, regions, and symbols. The symbol category is further divided into parallel marks, equal-length marks, right-angle marks, and length-label marks. Additional statistical details about visual keypoints $VK$, and length statistics for the problem text \(T_{p}\), the solution text \(T_{s}\), and the visual keypoint components \(vk_{n}\) are provided in \Cref{tab:data_analysis}.


\section{Task Definition}
We propose two tasks to evaluate a model’s \taskname capability, as illustrated in \Cref{fig:task_overview}.
The first task requires (1) identifying visual keypoints useful for subsequent explanation and (2) generating explanations that explicitly reference them.
For robust evaluation, we structured the tasks to isolate perceptual and reasoning subskills.
Although the design abstracts away some real-world complexity, this two-stage setup still provides a clear and measurable step toward unified, open-ended reasoning.

\paragraph{Visual Keypoint Identification.} 
The first task evaluates the model’s ability to identify visual keypoints that are crucial for comprehension. 
Since a problem may have multiple valid solutions and the corresponding visual keypoints can vary, we provide the model with a solution summary ($T_{s}^{tldr}$) that anchors a single explanatory direction. To ensure that the evaluation focuses on keypoint identification rather than problem-solving, the correct answer is also provided. Additionally, because current models cannot reliably generate valid keypoints and open-ended scoring is ambiguous, we adopt a multiple-choice format for robust evaluation.

Given a problem image $I_p$, its text $T_p$, the correct answer, and the solution summary $T_s^{tldr}$, the model must select, from five candidate sets, the visual keypoints ($VK$) essential for understanding. The four distractor sets are constructed as follows: (1) $VK$ from a problem whose text is semantically similar to $T_p$; (2) $VK$ from a problem whose solution summary resembles $T_s$; (3) $VK$ from a problem whose own keypoints closely match the target $VK$ (4) $VK$ from a randomly selected problem. Text similarity was computed using Qwen3 Embedding~\cite{qwen3embedding}, and to ensure reliability, human annotators carefully reviewed and corrected any options exhibiting logical inconsistencies.

\paragraph{Keypoint-based Explanation Generation.} 
The second task evaluates whether the model can effectively generate explanatory text grounded in the appropriate visual keypoints.
As in the first task, we provide visual keypoints ($VK$) to guide the model toward a single reasoning path and the correct answer to focus evaluation on keypoint-aligned explanation generation rather than problem-solving. 

Given a problem consisting of an image $I_{p}$, text $T_{p}$, and problem’s answer, along with the visual keypoints $VK$, the model is required to produce a solution explanation $T_{s}$ that refers to the relevant visual elements.

\begin{figure}[!t]
    \includegraphics[width=1\linewidth]{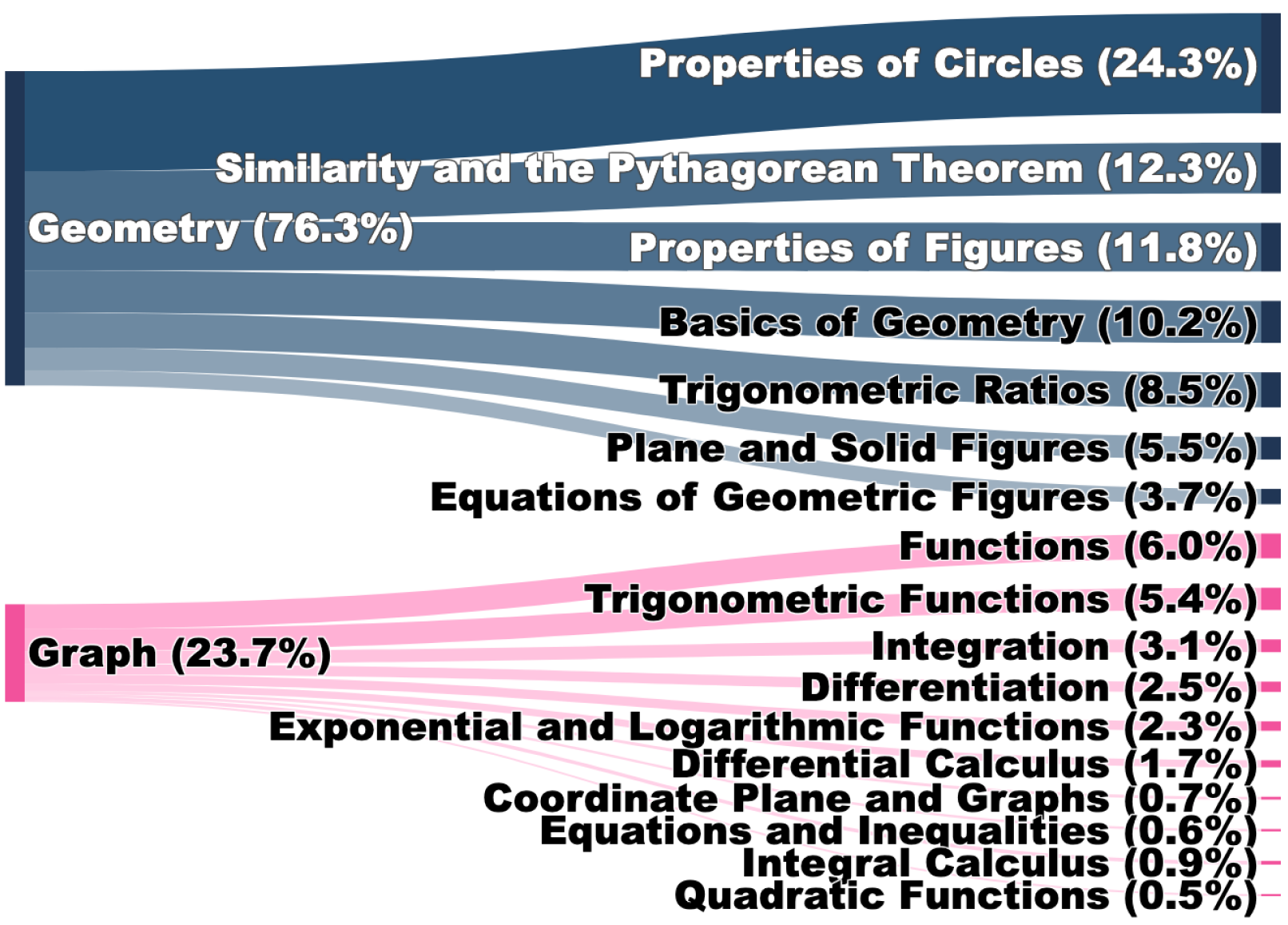}
    \caption{Topic coverage of geometry and graph across 17 chapters in the \benchmarkname benchmark.}
    \label{fig:data_analysis}
\end{figure}


\section{Experiments}
\paragraph{Models.}
We evaluate three categories of MLLMs:
(1) \textbf{generalist models}, including Molmo 7B~\cite{deitke2024molmo}, LLaVA-1.6 7B~\cite{liu2024improved}, Qwen2-VL 7B~\cite{wang2024qwen2}, and Qwen2.5-VL 7B \& 72B~\cite{bai2025qwen2};
(2) \textbf{math-specialized models}, including Math-PUMA 7B~\cite{zhuang2024math}, URSA 8B~\cite{luo2025ursa}, and Math-LLaVA 13B~\cite{shi-etal-2024-math}; and
(3) \textbf{proprietary models}, GPT-4o~\cite{achiam2023gpt} and Gemini 2.0 Flash~\cite{google2024gemini2}.
Details of the experimental setup and prompts are provided in the Appendix.

\subsection{Toy: Solution Recognition}
The \taskname tasks are designed to evaluate specific abilities rather than general problem-solving skills.
To examine whether the model genuinely understands how to solve problems, we first perform a preliminary study on \benchmarkname prior to the multimodal explanation task.

\paragraph{Metrics.}
\benchmarkname consists of both multiple-choice and short-answer problems. We report accuracy following the MathVista evaluation protocol \cite{lu2023mathvista}.

\paragraph{Results.}
Table~\ref{tab:task0} shows the accuracy of problem-solving. The 7B generalist baseline struggles, while the 72B model performs second best. Somewhat unexpectedly, the math-specialized models perform worse than the generalist models, likely due to hindered instruction-following capabilities. Among the proprietary baselines, GPT-4o struggled similarly to open-source models, whereas Gemini achieved the best performance overall. These results indicate that most MLLMs struggle to recognize the correct solution on \benchmarkname, even before performing the multimodal explanation task.

\begin{table}[t!]
\centering
\begin{tabular}{l|c|ccc}
\toprule
             & \multicolumn{1}{l|}{} & \multicolumn{3}{c}{Problem-Solving (Acc)}        \\ \midrule
Model        & Params               & Geometric      & Graph          & Overall        \\ \midrule
Molmo        & 7B                    & 0.248          & 0.194          & 0.235          \\
LLaVA-1.6    & 7B                    & 0.147          & 0.127          & 0.142          \\
Qwen2-VL     & 7B                    & 0.274          & 0.215          & 0.260          \\
Qwen2.5-VL   & 7B                    & 0.316          & 0.224          & 0.294          \\
Qwen2.5-VL   & 72B                   & {\underline{0.430}}    & \textbf{0.300} & {\underline{0.399}}    \\ \midrule
Math-PUMA    & 7B                    & 0.258          & 0.194          & 0.243          \\
URSA         & 8B                    & 0.055          & 0.068          & 0.058          \\
Math-LLaVA   & 13B                   & 0.202          & 0.152          & 0.190          \\ \midrule
GPT-4o       & -                     & 0.274          & 0.211          & 0.259          \\
Gemini 2.0 F & -                     & \textbf{0.481} & {\underline {0.291}}    & \textbf{0.436}
\\ \bottomrule
\end{tabular}
\caption{Experimental results on the \textit{Solution Recognition} toy task from \benchmarkname. Models are grouped into three categories: \textbf{generalist models} (top), \textbf{math-specialized models} (middle), and \textbf{proprietary models} (bottom). The best scores are in \textbf{bold}, and the second-best scores are \underline{underlined}.}
\label{tab:task0}
\end{table}

\subsection{Visual Keypoint Identification}
\paragraph{Metrics.}
We evaluate baseline performance using accuracy in a multiple-choice setting.

\paragraph{Results.}
\Cref{tab:task1} summarizes performance on the visual keypoint identification task. The 7B generalist models struggle, while the 72B model remains the second-best performer. In contrast, math-specialized models perform near chance level (Acc = 0.20), indicating severe difficulty in identifying visual cues. Among proprietary models, Gemini achieves the highest performance. Overall, most models struggle to identify visual keypoints even with access to the solution, though proprietary ones perform relatively better.

Since visual keypoint identification was evaluated under the assumption that models can already perform problem-solving, \Cref{tab:ps_vs_vki} reports success rates for each task and for both together. Only 23\%, 10\%, and 4\% of proprietary, generalist, and math-specialized models succeed on both.
This result highlights that real educational use remains challenging and that improving problem-solving ability is essential alongside visual keypoint identification.

\begin{table}[t!]
\centering
\begin{tabular}{l|c|ccc}
\toprule
             & \multicolumn{1}{l|}{} & \multicolumn{3}{c}{VK Identification (Acc)}      \\ \midrule
Model        & Params               & Geometric      & Graph          & Overall        \\ \midrule
Molmo        & 7B                    & 0.253          & 0.312          & 0.267          \\
LLaVA-1.6    & 7B                    & 0.260          & 0.283          & 0.265          \\
Qwen2-VL     & 7B                    & 0.273          & 0.371          & 0.296          \\
Qwen2.5-VL   & 7B                    & 0.363          & 0.532          & 0.403          \\
Qwen2.5-VL   & 72B                   & {\underline {0.486}}    & {\underline {0.696}}    & {\underline {0.536}}    \\ \midrule
Math-PUMA    & 7B                    & 0.194          & 0.219          & 0.200            \\
URSA         & 8B                    & 0.028          & 0.034          & 0.029          \\
Math-LLaVA   & 13B                   & 0.218          & 0.215          & 0.217          \\ \midrule
GPT-4o       & -                     & 0.418          & 0.646          & 0.472          \\
Gemini 2.0 F & -                     & \textbf{0.529} & \textbf{0.726} & \textbf{0.576}
\\ \bottomrule
\end{tabular}
\caption{Experimental results for the \textit{Visual Keypoint Identification} task on \benchmarkname, where models are evaluated on their ability to select the correct keypoints from multiple-choices.}
\label{tab:task1}
\end{table}

\begin{table}[t!]
\centering
\begin{tabular}{l|ccc}
\toprule
Model                            & PS Only & VKI Only & PS $\cap$ VKI \\ \midrule
$\text{Qwen2.5-VL}^{\text{7B}}$  & 18.9\%  & 29.8\%   & 10.5\%        \\
Math-PUMA                        & 19.7\%  & 15.5\%   & 4.6\%         \\
Gemini 2.0 F                     & 20.5\%  & 34.5\%   & 23.1\%     
\\ \bottomrule
\end{tabular}
\caption{Proportion (\%) of cases where each model succeeds only on Problem-Solving (PS), only on Visual Keypoint Identification (VKI), or on both (PS $\cap$ VKI)}
\label{tab:ps_vs_vki}
\end{table}

\begin{figure*}[!t]
    \includegraphics[width=1\linewidth]{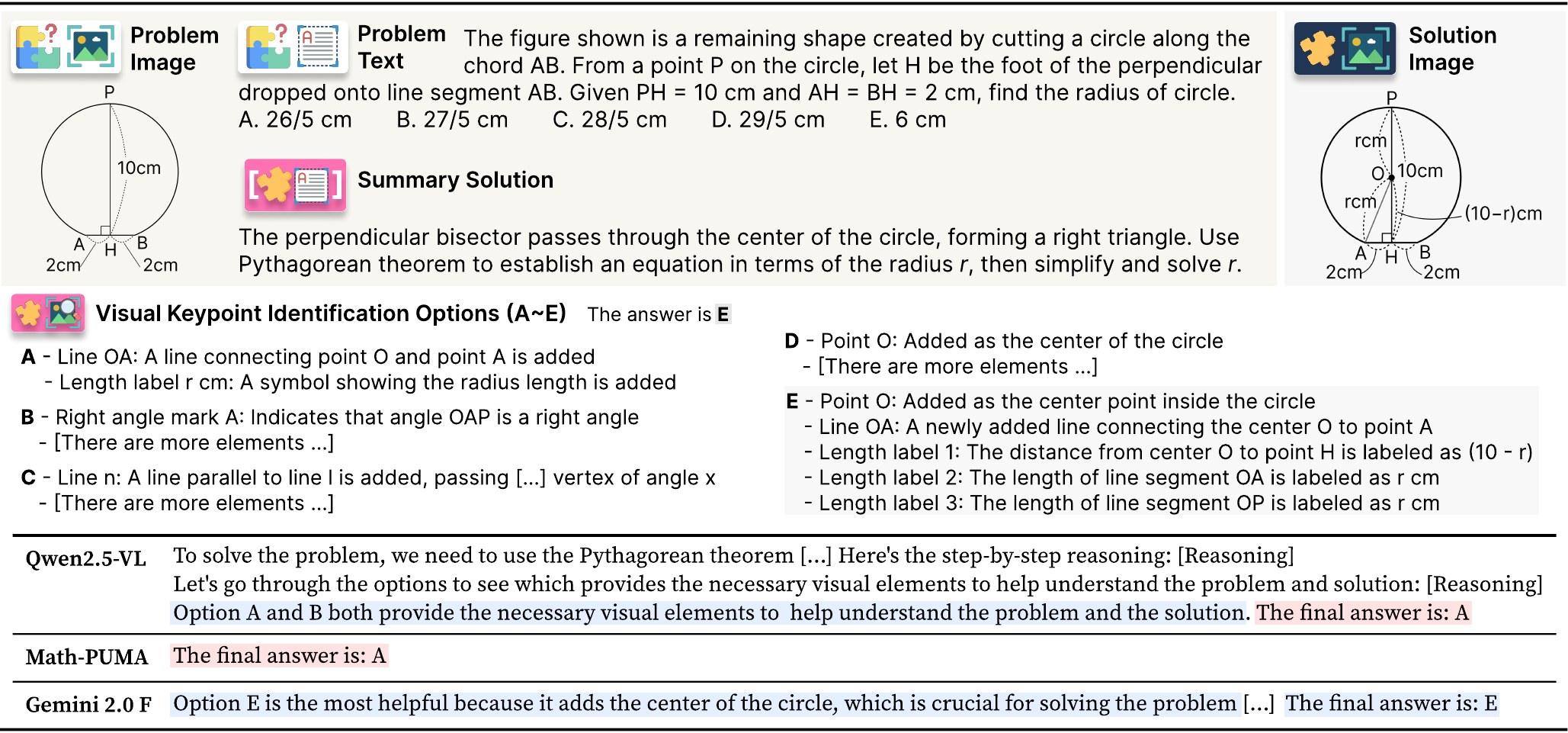}
    \caption{Examples of reasoning processes and final predictions produced by Qwen2.5-VL 7B, Math-PUMA, and Gemini 2.0 Flash on the \textit{Visual Keypoint Identification} task. Qwen2.5-VL demonstrates task understanding and reasoning but produces an incorrect answer, Math-PUMA lacks both, while Gemini 2.0 Flash demonstrates both and produces the correct answer.}
    \label{fig:qualitative1}
\end{figure*}

\subsection{Keypoint-based Explanation Generation}
\paragraph{Metrics.}
We evaluate the quality of the explanation using three criteria:  
(1) \textbf{Correctness} – whether the model’s reasoning is logically sound and leads to a valid solution;  
(2) \textbf{Fidelity} – whether the explanation aligns with the reasoning and intent of the reference, regardless of surface form;  
(3) \textbf{Referencing} – whether the explanation refers to the same key visual components (e.g. points, lines, etc) as the reference.
Each criterion is rated on a 5-point Likert scale. We report results from both human evaluators~\cite{zheng2023judging} and an LLM-based evaluator using GPT-4o~\cite{achiam2023gpt}. In addition, we report text similarity metrics, including BLEU, ROUGE, METEOR, and BERTScore~\cite{papineni2002bleu, lin2004rouge, banerjee2005meteor, zhang2019bertscore}, with further details provided in the Appendix.

\begin{table}[t!]
\centering
\begin{tabular}{l|ccc}
\toprule
Model                            & Correctness    & Fidelity       & Referencing    \\ \midrule
Molmo                            & 2.362          & 1.906          & 2.545          \\
LLaVA-1.6                        & 1.815          & 1.545          & 1.970          \\
Qwen2-VL                         & 1.844          & 1.623          & 1.898          \\
$\text{Qwen2.5-VL}^{\text{7B}}$  & 3.005          & 2.375          & 3.132          \\
$\text{Qwen2.5-VL}^{\text{72B}}$ & 3.397          & 3.048          & 3.533          \\ \midrule
Math-PUMA                        & 1.770          & 1.609          & 1.581          \\
URSA                             & 1.148          & 1.140          & 1.155          \\
Math-LLaVA                       & 2.100          & 1.360          & 1.286          \\ \midrule
GPT-4o                           & {\underline {3.784}}    & {\underline {3.153}}    & {\underline {3.892}}    \\
Gemini 2.0 F                     & \textbf{3.849} & \textbf{3.489} & \textbf{4.103}
\\ \bottomrule
\end{tabular}
\caption{LLM-based evaluation results for the \textit{Keypoint-based Explanation Generation} task on \benchmarkname, rated on a 1-5 Likert scale across three criteria: (1) Correctness, assessing logical validity; (2) Fidelity, measuring alignment with the intent of the reference explanation; and (3) Referencing, evaluating the appropriate use of key visual elements.}
\label{tab:task2-1}
\end{table}

\paragraph{Results.}
\Cref{tab:task2-1} presents the results of the explanation generation task. While most models achieve reasonable Correctness, many fail to follow the intended reasoning path (Fidelity) or reference the given keypoints (Referencing). Generalist models struggle overall, though the Qwen2.5-VL series shows size-dependent improvement. Math-specialized models still fail to produce coherent or instruction-following explanations. In contrast, proprietary models achieve the highest scores, demonstrating stronger abilities in generating well-grounded explanations.

As shown in \Cref{tab:task2-2}, human evaluation exhibits strong correlations with LLM judgments, as indicated by the Spearman coefficients~\cite{zar2005spearman} (0.770 for Correctness, 0.783 for Fidelity, 0.788 for Referencing; all p$<$0.05). These results show that although most open-source models still struggle, proprietary models and more recent generalist models can generate appropriate explanations.

\section{Analyses}
\subsection{Qualitative Analysis}
To analyze how the three categories of models differ in their outputs, we examined the results from representative models in each category: Qwen2.5-VL 7B (generalist), Math-PUMA (specialist), and Gemini 2.0 Flash (proprietary).

\begin{table}[t!]
\centering
\begin{tabular}{l|ccc}
\toprule
Model                           & Correctness    & Fidelity       & Referencing    \\ \midrule
$\text{Qwen2.5-VL}^{\text{7B}}$ & \underline{2.610}          & \underline{2.541}          & \underline{2.610}          \\
Math-PUMA                       & 1.041          & 1.041          & 1.037          \\
Gemini 2.0 F                    & \textbf{4.423} & \textbf{4.171} & \textbf{4.256}

\\ \bottomrule
\end{tabular}
\caption{Human evaluation results for the \textit{Keypoint-based Explanation Generation} task, rated on a 1–5 Likert scale.}
\label{tab:task2-2}
\end{table}

\begin{figure}[!t]
    \includegraphics[width=1\linewidth]{fig_failure_category.png}
    \caption{
    Error analysis for \textit{Visual Keypoint Identification}: (1) cases where the correct element was chosen but referenced incorrectly, (2) choices containing more keypoints than required, and (3) choices containing fewer keypoints than needed.}
    \label{fig:failure_category}
\end{figure}

\paragraph{Visual Keypoint Identification.} 
\Cref{fig:qualitative1} shows visual keypoint identification examples from three model categories. Qwen2.5-VL attempts to reason about the most informative keypoints but still selects the wrong option. Math-PUMA shows neither coherent reasoning nor a correct answer. In contrast, Gemini 2.0 Flash correctly interprets the instruction, analyzes the candidates, and chooses the most appropriate keypoints. Overall, similar patterns were consistently observed across examples.

To gain a finer-grained understanding of the models, we analyze their output behaviors on a 10\% subset of the dataset, as shown in \Cref{fig:failure_category}. We categorize all incorrect outputs into three types: (1) electing the correct elements but providing incorrect descriptions, (2) selecting options that contain extra elements, and (3) selecting options with missing required elements. The three models exhibit similar rates for incorrect descriptions and extra elements but differ substantially in missing elements: Math-PUMA shows the highest rate (38\%), followed by Qwen2.5-VL (20\%). In contrast, Gemini 2.0 Flash achieves the highest accuracy, while Qwen2.5-VL performs moderately, and Math-PUMA remains the least reliable among the three.

\begin{figure}[!t]
    \includegraphics[width=1\linewidth]{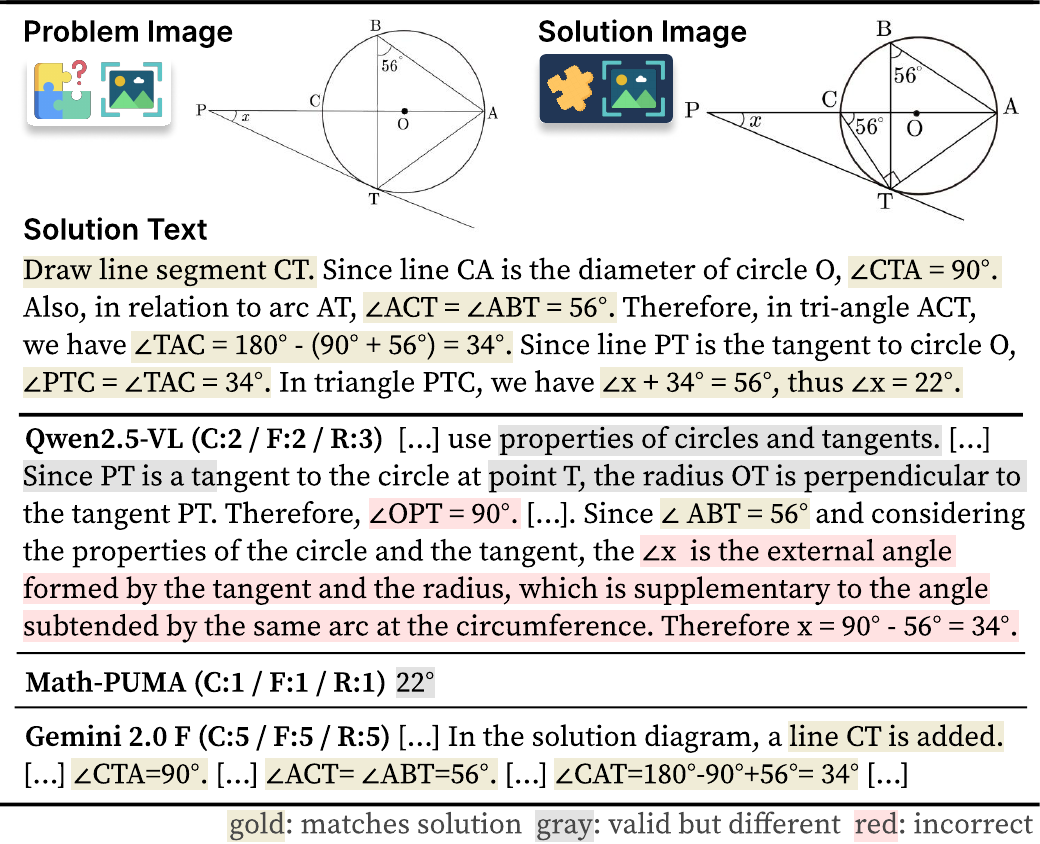}
    \caption{
    Examples of explanations from three models on the \textit{Keypoint-based Explanation Generation} task, along with their evaluated scores in Correctness (C), Fidelity (F), and Referencing (R). Qwen2.5-VL starts with valid yet unaligned explanations, which soon become incorrect; Math-PUMA generates no explanation; Gemini 2.0 Flash generates solution-aligned explanations.}

    \label{fig:qualitative2}
\end{figure}

\paragraph{Keypoint-based Explanation Generation.}
\Cref{fig:qualitative2} presents explanation examples from three model categories. 
In this example, Qwen2.5-VL scored 2 in Correctness, 2 in Fidelity, and 3 in Referencing. Despite being provided with keypoints, its reasoning diverges from the reference and only partially aligns with the solution image, ultimately leading to an incorrect interpretation.
Math-PUMA received a score of 1 across all metrics, as it only returned the final answer without any supporting explanation.
In contrast, Gemini scored 5 across all metrics, generating an explanation that mirrored the reference reasoning and consistently referred to the correct key visual elements.
Similar to the experimental results in \Cref{fig:failure_category}, proprietary models generate coherent explanations, whereas open-source models struggle, with math-specialized ones showing almost no reasoning capability.

\begin{figure}[!t]
    \includegraphics[width=1\linewidth]{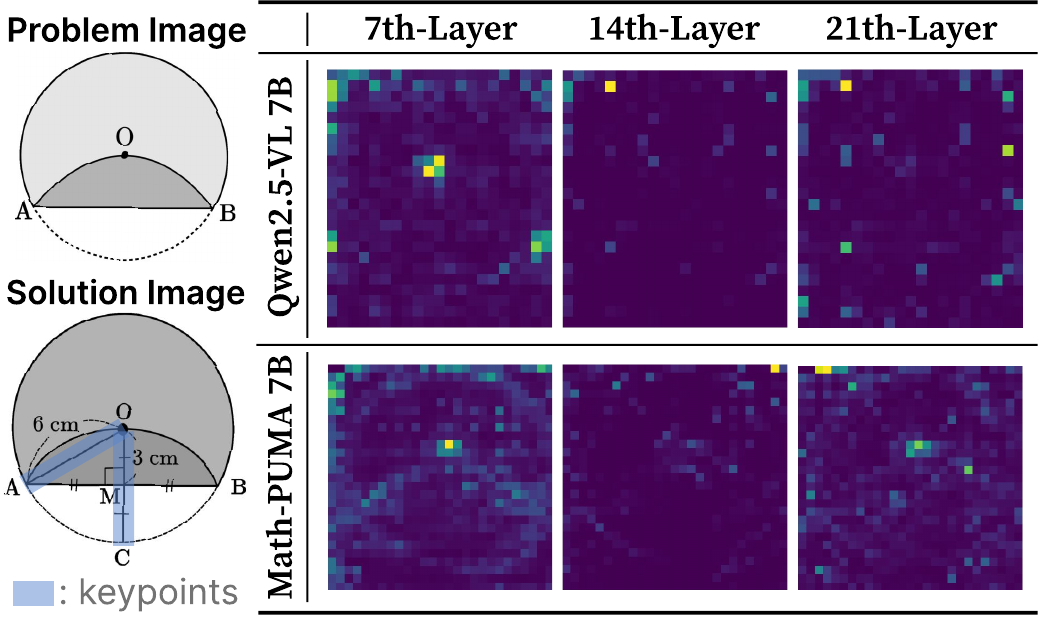}
    \caption{Attention maps from various layers of open-source MLLMs on the \textit{Visual Keypoint Identification} task. While the models attend visually to the problem image, they fail to focus on the keypoints that are most relevant for explanation.}
    \label{fig:attmap}
\end{figure}

\subsection{Do MLLMs Attend to Visual Keypoints?}
To analyze how current open-source models fail to visually recognize keypoints, we examined the attention maps of both generalist and math-specialized models on the Visual Keypoint Identification task. \Cref{fig:attmap} illustrates the attention patterns, showing that both models tend to attend well to both global and local regions of the input image. However, despite the explicit inclusion of visual keypoints in the options and summary solution, neither model effectively focuses on the relevant visual regions. A similar pattern is observed in the Keypoint-based Explanation Generation task. While current MLLMs are capable of attending to general visual content~\cite{zhang2025mllms}, our analysis suggests that they still lack robust mathematical visual grounding ability. Improving visual grounding in mathematical contexts will likely be essential for future models to perform well on the \taskname task.


\section{Conclusion}
We introduce the \textit{\taskname} task and the \benchmarkname benchmark, which assess multimodal mathematics‐teaching capabilities through two complementary subtasks: identifying essential visual keypoints and generating explanations grounded in those keypoints. 
Our experimental results demonstrate that current models struggle with both problem-solving and visual keypoint identification, and that this performance gap becomes more pronounced for open-source models in the explanation generation task.
Through our analysis, we find that this limitation stems from the lack of math-specific visual grounding and robust visually grounded reasoning. 
Enhancing these two abilities will be essential for applying MLLMs effectively in educational settings.

\section*{Ethics Statement}
In this paper, we introduce the \benchmarkname benchmark, curated from an in-house mathematics education platform\footnote{\url{https://mathpresso.com/en}}. All materials were reviewed to ensure full compliance with copyright and data usage regulations.
Data annotation was conducted by bilingual annotators holding undergraduate degrees in relevant fields, ensuring adequate mathematical and linguistic proficiency.

\section*{Acknowledgments}
This work was partly supported by an Institute of Information \& communications Technology Planning \& Evaluation (IITP) grant funded by the Korean Government (MSIT) (No.~RS-2021-II211343, Artificial Intelligence Graduate School Program (Seoul National University), No.RS-2025-02263598, Development of Self-Evolving Embodied AGI Platform Technology through Real-World Experience), the National Research Foundation of Korea(NRF) grant funded by the Korea government(MSIT)(RS-2024-00354218, RS-2024-00353125). We express special thanks to KAIT GPU project. The ICT at Seoul National University provides research facilities for this study. 

\bibliography{aaai2026}

\clearpage
\appendix
\setcounter{page}{1}
\renewcommand{\thesection}{\Alph{section}}
\setcounter{section}{0} 
\renewcommand{\thetable}{\Alph{table}}
\setcounter{table}{0}
\renewcommand{\thefigure}{\Alph{figure}}
\setcounter{figure}{0}

\definecolor{codebg}{rgb}{0.95, 0.95, 0.95}  
\lstdefinestyle{mystyle}{
    backgroundcolor=\color{codebg},   
    basicstyle=\ttfamily\small,       
    frame=single,                     
    breaklines=true,                   
    captionpos=b,                      
    keywordstyle=\bfseries,            
    commentstyle=\color{gray},         
    numbersep=5pt,                     
    xleftmargin=5pt, xrightmargin=5pt  
}

\twocolumn[
    \centering
    \LARGE 
    \textbf{Explain with Visual Keypoints Like a Real Mentor!\\
    A Benchmark for Multimodal Solution Explanation}\\
    \vspace{0.5em}Technical Appendix\\
    \vspace{2em}
]


\section{Experiment Details}
In this study, experiments were conducted using the LMMs-eval repository~\footnote{\url{https://github.com/EvolvingLMMs-Lab/lmms-eval}}. This repository provides a comprehensive framework for evaluating multi-modal models across various tasks.

\subsection{Computational Resources}
For closed-source models, we used the OpenAI API and Gemini Developer API to infer the output of GPT-4o and Gemini 2.0. For open-source models such as Math-PUMA, URSA, MathLLaVA, LLaVA, Qwen-2-VL, Qwen-2.5-VL, and Molmo, inference was performed using a NVIDIA A6000 48GB GPU. While the exact inference speed varies depending on the task, model, and lengths of prompts and responses, a query takes about 50 seconds to be answered.

\subsection{Evaluated Models}
When conducting experiments with open-source multimodal models, we leveraged the official implementation codes in conjunction with publicly available weights from the Huggingface Hub~\footnote{\url{https://huggingface.co/models}}. The following model parameters were used for each model:

\begin{itemize}
    \item \footnotesize\textbf{Math-PUMA:} {\texttt{Math-PUMA/Math-PUMA\_Qwen2VL-7B}}
    \item \footnotesize\textbf{URSA:} {\texttt{URSA-MATH/URSA-RM-8B}}
    \item \footnotesize\textbf{Math-LLaVA:} {\texttt{Zhiqiang007/Math-LLaVA}}
    \item \footnotesize\textbf{llava-1.6:} {\texttt{llava-hf/llava-v1.6-mistral-7b-hf}}
    \item \footnotesize\textbf{Qwen2-VL-7B:} {\texttt{Qwen/Qwen2-VL-7B-Instruct}}
    \item \footnotesize\textbf{Qwen2.5-VL-7B:} {\texttt{Qwen/Qwen2.5-VL-7B-Instruct}}
    \item \footnotesize\textbf{Qwen2.5-VL-72B:} {\texttt{Qwen/Qwen2.5-VL-72B-Instruct}}
    \item \footnotesize\textbf{Molmo:} {\texttt{allenai/Molmo-7B-D-0924}}
\end{itemize}

These models were evaluated in our benchmark, which included tasks designed to assess both visual understanding and textual explanation capabilities. The selection of models spans a range of architectures and performance levels, providing insights into current advancements in multi-modal learning.

\subsection{LLM Evaluation}
In all LLM-based evaluations, we used the \texttt{gpt-4o-2024-08-06} endpoint. For the Keypoint-based Explanation Generation task, we compared the rankings obtained when Math-PUMA and GPT outputs were evaluated separately by Gemini and GPT. The resulting Kendall’s $\tau$ values were 0.90 for Correctness, 0.84 for Fidelity, and 0.92 for Referencing. Although evaluation bias is often a concern when an LLM assesses models from the same family, the high agreement between the GPT-judge and Gemini-judge indicates that no substantial bias is present.

\subsection{Human Evaluation}
Three evaluators, all holding a bachelor’s or master’s degree in engineering, are assessing AI model outputs for 80 problems. Specifically, they are evaluating the results produced by the Math-PUMA, Qwen2.5-VL, and Gemini 2.0 Flash models, which represent the math-specialized, generalist, and proprietary models categories. Each criterion is being rated on a five-point Likert scale. LLM–human agreement shows strong correlations, as indicated by the Spearman coefficients (0.770 for Correctness, 0.783 for Fidelity, and 0.788 for Referencing; all $p<0.05$). Human–Human agreement, measured by Krippendorff’s $\alpha$, reached 0.696 for Correctness, 0.571 for Fidelity, and 0.612 for Referencing.

\subsection{Automatic Metrics Evaluation}
We report additional evaluation results for Keypoint-based Explanation Generation task using automatic metrics, including BLEU, ROUGE, METEOR, and BERTScore. The detaileed scores can be found in \Cref{tab:apdx_explanation_generation_task_evaluation}.

\section{Benchmark Details}
Our \benchmarkname benchmark consists of a total of 17 chapters and 33 sections as shown in \Cref{tab:apdx_benchmark_detail}. During dataset validation, we also reviewed the options related to visual keypoint identification and confirmed their consistency and reliability.

\section{Ablation on Solution Summary Anchoring}
\begin{figure*}[!t]
    \includegraphics[width=1\linewidth]{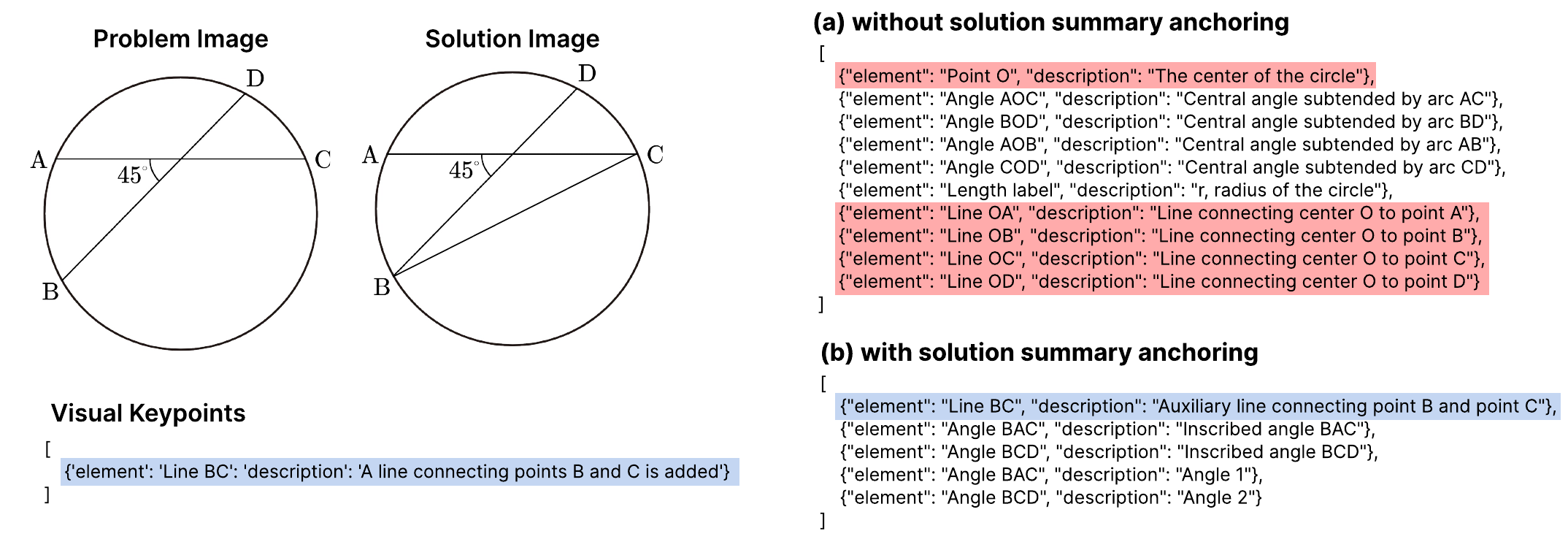}
    \caption{Comparison of explanations generated with and without the solution summary ($T_{s}^{tldr}$). Without anchoring, models often drift to alternative reasoning paths or irrelevant keypoints. (a) Without $T_{s}^{tldr}$; (b) With $T_{s}^{tldr}$.}
    \label{fig:apdx_tldr_comparison}
\end{figure*}

Since a problem may have multiple valid solutions and the corresponding visual keypoints can vary, we provide the model with a solution summary ($T_{s}^{tldr}$) that anchors a single explanatory direction. 
To examine the effect of this anchoring, we analyze the qualitative differences when the summary is not provided in \Cref{fig:apdx_tldr_comparison}. 
As shown, without anchoring, the model’s explanations often drift toward alternative reasoning paths or focus on irrelevant keypoints.

\begin{table*}[!t]
\centering
\begin{tabular}{l|c|ccccc}
\toprule
Model & Params & BLEU-2 & BLEU-4 & ROUGE-L & METEOR & BERTScore \\ 
\midrule
Molmo & 7B & {\underline{0.158}} & 0.067 & 0.187 & 0.310 & {\underline{0.842}} \\
LLaVA-1.6 & 7B & 0.130 & 0.059 & 0.176 & 0.287 & 0.835 \\
Qwen2-VL & 7B & \textbf{0.176} & \textbf{0.087} & \textbf{0.237} & 0.288 & \textbf{0.854} \\
Qwen2.5-VL & 7B & 0.099 & 0.038 & 0.193 & 0.284 & 0.819 \\
Qwen2.5-VL & 72B & 0.097 & 0.043 & 0.190 & {\underline{0.316}} & 0.821 \\ 
\midrule
Math-PUMA & 7B & 0.006 & 0.002 & 0.119 & 0.058 & 0.818 \\
URSA & 8B & 0.020 & 0.006 & 0.079 & 0.075 & 0.735 \\
Math-LLaVA & 13B & 0.112 & 0.057 & 0.147 & 0.252 & 0.817 \\ 
\midrule
Gemini 2.0 Flash & - & 0.149 & {\underline{0.083}} & {\underline{0.210}} & \textbf{0.367} & {\underline{0.842}} \\
GPT-4o & - & 0.095 & 0.045 & 0.161 & 0.301 & 0.815 \\ 
\bottomrule
\end{tabular}
\caption{Experimental results of automated evaluation metrics for the \textit{Keypoint-based Explanation Generation} task on \benchmarkname.}
\label{tab:apdx_explanation_generation_task_evaluation}
\end{table*}

\section{Prompts}
This section compiles all the prompts used in our experiments. The prompts shown in \Cref{fig:prompt_visual_keypoint_identification}, \Cref{fig:prompt_explanation_generation}, and \Cref{fig:prompt_problem_solving} are used to generate model outputs for the Solution Recognition toy task, Visual Keypoint Identification, and Keypoint-based Explanation Generation tasks, respectively. For the Keypoint-based Explanation Generation task, model responses are evaluated using the prompts in \Cref{fig:prompt_Evaluation_explanatory_text_generation}.

\begin{table*}[!t]
\centering
\begin{tabular}{@{}l|l@{}}
\toprule
\textbf{Chapter Title} & \textbf{Section Title} \\ \midrule
\multirow{2}{*}{Basics of Geometry} & Basic Geometric Figures \\ 
 & Construction and Congruence \\ \midrule
Coordinate Plane and Graphs & Coordinate Plane and Graphs \\ \midrule
Differential Calculus & Differentiation of Various Functions \\ \midrule
\multirow{2}{*}{Differentiation} & Derivative and Derivative Function \\ 
 & Applications of Derivatives \\ \midrule
Equations and Inequalities & Quadratic Equations and Functions \\ \midrule
\multirow{4}{*}{Equations of Geometric Figures} & Transformations of Figures \\ 
 & Equation of a Circle \\ 
 & Coordinate Plane \\ 
 & Equations of Straight Lines \\ \midrule
Exponential and Logarithmic Functions & Exponential and Logarithmic Functions \\ \midrule
\multirow{4}{*}{Functions} & Linear Functions and Their Graphs \\ 
 & Functions \\ 
 & Rational and Irrational Functions \\ 
 & Relationship Between Linear Functions and Equations \\ \midrule
\multirow{2}{*}{Integral Calculus} & Applications of Definite Integrals \\ 
 & Various Integration Techniques \\ \midrule
\multirow{2}{*}{Integration} & Indefinite and Definite Integrals \\ 
 & Applications of Definite Integrals \\ \midrule
\multirow{2}{*}{Plane and Solid Figures} & Properties of Solid Figures \\ 
 & Properties of Plane Figures \\ \midrule
\multirow{2}{*}{Properties of Circles} & Circle and Line \\ 
 & Inscribed Angles \\ \midrule
\multirow{2}{*}{Properties of Figures} & Properties of Quadrilaterals \\ 
 & Properties of Triangles \\ \midrule
Quadratic Functions & Graph of the Quadratic Function $y = ax^2 + bx + c$ \\ \midrule
\multirow{3}{*}{Similarity and the Pythagorean Theorem} & Pythagorean Theorem \\ 
 & Similarity of Figures \\ 
 & Applications of Similarity \\ \midrule
\multirow{2}{*}{Trigonometric Functions} & Meaning and Graphs of Trigonometric Functions \\ 
 & Law of Sines and Law of Cosines \\ \midrule
\multirow{2}{*}{Trigonometric Ratios} & Trigonometric Ratios \\ 
 & Applications of Trigonometric Ratios \\
\bottomrule
\end{tabular}
\caption{Overview of the 17 chapters and their corresponding 33 sections covered in the dataset.}
\label{tab:apdx_benchmark_detail}
\end{table*}

\begin{figure}[t]
\centering
\begin{tcolorbox}[
    colback=white, %
    colframe=gray, %
    arc=4mm, %
    fontupper=\small
]
You should choose a set of visual elements from the multiple-choice options (A, B, C, D, or E) that best reflect how a teacher would visually guide a student to understand and solve the problem.\\\\
Problem: \\\textbf{As shown in the figure, there are 5 points: A, B, C, D, and E. When selecting two points among them to form straight lines and rays, let the number of straight lines be a and rays be b. Find the value of a + b.} 
\\\\Answer: \textbf{19}

The solution process for the problem is as follows:
\\\textbf{\\Count the possible straight lines formed by selecting pairs of points, then count the rays formed by considering directionality. Add both counts to find the total.}
\\\textbf{\\A.- line AE: A line connecting point A and E}
\\\textbf{\\B.- Symbol a: Represents the line extending from the upper left to the lower right\\- Symbol b: Represents the line extending from the lower left to the upper right\\- Symbol c: Represents the horizontal line}
\\\textbf{\\C.- Line AB: A line extended from side AB of the hexagon\\- Line BC: A line extended from side BC of the hexagon\\- Line CD: A line extended from side CD of the hexagon\\- Line DE: A line extended from side DE of the hexagon\\- Line EF: A line extended from side EF of the hexagon\\- Line FA: A line extended from side FA of the hexagon}
\\\textbf{\\D.- Auxiliary line BD: A line segment connecting point B and point D is added}
\\\textbf{\\E.- Line AE: A straight line connecting points A and E\\- Line BE: A straight line connecting points B and E\\- Line CE: A straight line connecting points C and E\\- Line DE: A straight line connecting points D and E}\\\\
Based on this reasoning guidance, select **only one** of the option (A, B, C, D, or E) whose visual elements would be most helpful for students in understanding the problem and its solution.\\\\
Think carefully about how the selected visual elements support the reasoning process. You may briefly explain your thinking, but your response **must end** with the following format:\\
The final answer is: {A, B, C, D, or E}\\
IMPORTANT!! Your final response must END with the format.

\end{tcolorbox}
\caption{Prompt used for the \textit{Visual Keypoint Identification} task in the \benchmarkname benchmark. Prompt inputs are \textbf{boldfaced}.}
\label{fig:prompt_visual_keypoint_identification}
\end{figure}

\begin{figure}[b]
\centering
\begin{tcolorbox}[
    colback=white, %
    colframe=gray, %
    arc=4mm, %
    fontupper=\small %
]

Q: \textbf{As shown in the figure, there are 5 points: A, B, C, D, and E. When selecting two points among them to form straight lines and rays, let the number of straight lines be a and rays be b. Find the value of a + b.} \\
\#\#\# Answer: \textbf{19} \#\#\#\\

\#\#\# Difference between the original image and the solution image \#\#\#\\
\textbf{Line AE: A straight line connecting points A and E}\\
\textbf{Line BE: A straight line connecting points B and E}\\
\textbf{Line CE: A straight line connecting points C and E}\\
\textbf{Line DE: A straight line connecting points D and E}\\

You are a math teacher helping students understand how to solve problems clearly and effectively.\\
Given a problem description, problem image and a list of key elements introduced or highlighted in the solution image, write an educational explanation that helps students.\\
Additionally, this problem is a problem of \textbf{Functions/Linear Functions and Their Graphs} chapter. You should explain the problem in the context of the chapter and section.\\
Make sure to reference both the original components from the problem image and any new annotations, highlights, or added elements from the solution image to enhance understanding.\\

\#\#\# OUTPUT Example:\\
\{\\
\noindent\hspace*{1em}solution\_text: \\
\}
\end{tcolorbox}
\caption{Prompt used for the \textit{Keypoint-based Explanation Generation} task in the \benchmarkname benchmark. It is designed to generate educationally effective explanations for the given math problem. Prompt inputs are \textbf{boldfaced}.}
\label{fig:prompt_explanation_generation}
\end{figure}

\begin{figure}[b!]
\centering
\begin{tcolorbox}[
    colback=white, %
    colframe=gray, %
    arc=4mm, %
    fontupper=\small %
]

You are a math solver. For the problem below, **your task is ONLY to output the final answer** in one line.\\
**Do NOT provide any explanation, steps, or clarification. Just write the answer.**\\\\
Problem: \textbf{As shown in the figure, there are 5 points: A, B, C, D, and E. When selecting two points among them to form straight lines and rays, let the number of straight lines be a and rays be b. Find the value of a + b.}\\\\
Again, only return the final answer. Any additinal text will be considered incorrect.

\end{tcolorbox}
\caption{Prompt used for the \textit{Solution Recognition} toy task in the \benchmarkname benchmark. It is designed to generate an answer to the given math problem. Prompt inputs are \textbf{boldfaced}.}
\label{fig:prompt_problem_solving}
\end{figure}

\begin{figure*}[b]
\centering
\begin{tcolorbox}[
    colback=white, %
    colframe=gray, %
    arc=4mm %
]

You are evaluating the quality of an AI-generated explanation for a math problem involving geometry or graph-based reasoning.\newline

You will be given two texts:\newline

1. A reference explanation written by a human teacher.\newline
2. An AI-generated explanation written by a model.\newline

Your task is to compare the two explanations and assess how accurately and effectively the AI-generated explanation captures the key geometric concepts and reasoning presented in the reference.\newline

Please evaluate the model's explanation and provide four scores based on the criteria below:\newline

---\newline
\#\#\# Scoring Criteria\newline

1. Correctness  \newline
   - Does the reasoning presented by the model make sense and help solve the problem appropriately?  \newline
   - Rate on a Likert scale: **1, 2, 3, 4, or 5**\newline

2. Reference Alignment  \newline
   - Does the model follow the same logical reasoning and intent as the reference explanation, even if the wording differs?  \newline
   - Rate on a Likert scale: **1, 2, 3, 4, or 5**\newline

3. Use of Key Visual Elements  \newline
   - Does the AI explanation refer to the same critical visual components (e.g., points, lines, angles, shapes) as the reference?  \newline
   - Alternative terminology is acceptable if it clearly refers to the same element or serves the same purpose.  \newline
   - Rate on a Likert scale: **1, 2, 3, 4, or 5**\newline

---\newline
\#\#\# Output Format\newline

Important: Report your rating using the exact format below:\newline

Rating: [[x, y, z]]  \newline
— where `x` is your score for correctness, `y` for reference alignment, and `z` for use of visual elements.

\noindent\hspace*{2em}

\end{tcolorbox}
\caption{GPT evaluation prompt used to assess model outputs for the \textit{Keypoint-based Explanation Generation} task in \benchmarkname. Prompt inputs are \textbf{boldfaced}.}
\label{fig:prompt_Evaluation_explanatory_text_generation}
\end{figure*}
\end{document}